\documentclass[final,12pt,3p]{elsarticle}

\usepackage{amssymb}
\usepackage{listings}
\usepackage{amsmath}
\usepackage{subfig}
\usepackage{pdflscape}
\usepackage{array}
\usepackage{multirow}
\usepackage{lscape}

\usepackage{hyperref}
\hypersetup{
    colorlinks=true,
    linkcolor=blue,
    filecolor=magenta,      
    urlcolor=cyan,
}

\journal{arxiv}

\begin{document}

\begin{frontmatter}

\title{A Structured-Light Scanning Software for Rapid Geometry Acquisition}

\author[label1]{Qing Gu\corref{cor1}}
\ead{tsing.goo@gmail.com}

\author[label2]{Kyriakos Herakleous\corref{cor2}}
\ead{kyriakosv.2005@hotmail.com}

\author[label3]{Charalambos Poullis}
\ead{charalambos@poullis.org}
\address[label1, label2, label3]{Immersive and Creative Technologies Lab, Department of Computer Science and Software Engineering, Concordia University}

\cortext[cor1]{Lead researcher from v4 [CPU + GPU].}
\cortext[cor2]{Lead researcher up to and including v3.2.}

\begin{abstract}
Recently, there has been an increase in the demand of virtual 3D objects representing real-life objects. A plethora of methods and systems have already been proposed for the acquisition of the geometry of real-life objects ranging from those which employ active sensor technology, passive sensor technology or a combination of various techniques. 

In this paper we present the development of a 3D scanning system which is based on the principle of structured-light, without having particular requirements for specialized equipment. We discuss the intrinsic details and inherent difficulties of structured-light scanning techniques and present our solutions. Finally, we introduce our open-source scanning software system "3DUNDERWORLD-SLS" which implements the proposed techniques both in CPU and GPU. We have performed extensive testing with a wide range of models and report the results. Furthermore, we present a comprehensive evaluation of the system and a comparison with a high-end commercial 3D scanner.
\end{abstract}


\end{frontmatter}


\section{Introduction}
\label{sec:SLS}
In recent years, there is an increasing demand for 3D models and in particular 3D models representing/replicating real-world objects. Of the many available techniques (\cite{schuon2008high}, \cite{cui20103d})(\cite{bajracharya2012real},\cite{geiger2011stereoscan}), Structured-light Scanning (SLS)(\cite{gupta2011structured}) systems have emerged as the most cost-effective and accurate method to capture the 3D geometry and appearance of a real object. 

SLS systems employ active-sensors such as projectors and laser emitters to project light of a known structure (pattern). The scanning process involves the projection of a series of these known patterns on the object being scanned, while capturing the scene from one or more cameras. The projected pattern will be distorted according to the geometry of the object being scanned. The geometry of the object can then be computed by identifying correspondences between the pixels in the images captured by the camera(s). In order to be able to identify these pixel correspondences, the projected pattern is encoded such that every pixel in the projected pattern can be uniquely identified in the images captured by the cameras. This provides an efficient and very accurate method of mapping corresponding pixels between the images captured by the cameras. Finally, using this mapping between corresponding image pixels, it is possible to calculate their accurate 3D positions.

Many variants of SLS systems have already been proposed each one tailored to a particular task. Although the theory behind the SLS systems is well documented and understood, there are still many issues one has to consider when developing or using SLS systems, which are currently lacking documentation. In this paper, we present all the intrinsic details, limitations and solutions one has to consider when involved with the design, development and application of SLS systems. Moreover, we introduce the open-source scanning system 
"3DUNDERWORLD-SLS" and present the results of extensive testing.

The paper is organized as follows: Section \ref{sec:related_work} provides a brief overview of state-of-the-art in the area of 3D scanning, and Section \ref{sec:system_overview} gives an overview of the scanning system. The different schemes for the encoding of the patterns are presented in Section \ref{sec:encoding_of_patterns} and the calibration of the cameras is described in Section \ref{sec:calib}. Section \ref{sec:acquisition} describes the data acquisition. The captured images are decoded as explained in Section \ref{sec:decoding} and the reconstructed points are calculated as described in Section \ref{sec:reconstruction}. Section \ref{sec:point2Mesh} describes how the points can be finally converted to a mesh. In Section \ref{sec:results} we report on the results of our extensive testing and in Section \ref{sec:evaluation} we provide a comprehensive evaluation of the scanning system. 

\begin{figure*}[!ht]
 \centering
 \includegraphics[scale=1]{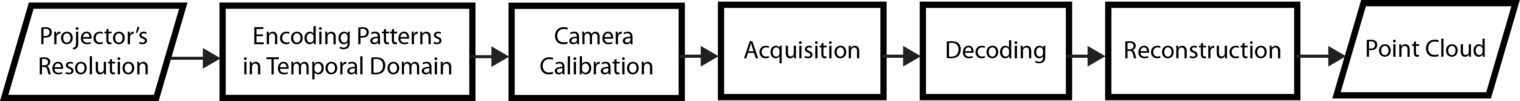}
 \caption{\label{fig:system_overview} System Overview}
\end{figure*}

\section{Related Work}
\label{sec:related_work}
A plethora of work has been done in the area of scanning and in particular structured-light scanning. Below we present a brief overview on the state-of-the-art in the area of scanning systems:

Microsoft's Kinect\cite{smisek20133d} is currently attracting the attention of researchers in the area. The device employs an infrared camera and pattern projector system to enable 3D vision in order to recognize people in the scene. Many uses of this device have already been proposed for fast 3D geometry acquisition, even for full body scanning as reported in \cite{cui20113d} \cite{tong2012scanning} for 3D avatar building and 3D human animation purposes. 

There are also uses of the device combined with other techniques for higher resolution results, such as the system proposed in \cite{somanath2013stereo+} which integrates traditional stereo matching with Kinect's information to take the advantages of both.
In \cite{fritsch2012photogrammetric} a multi-camera system for dense 3D Point cloud computation is proposed. The system uses 5 cameras and a Kinect device. In this case the Kinect device is not directly used for the 3D acquisition, but rather as a light source where the IR projector embeds features in the scene which the 4 infrared sensitive cameras are observing while the fifth is used for movement registration. Similarly, in \cite{rohith2009camera} a camera flash based projector is presented for use in stereo camera systems (with two cameras) to improve stereo matching.

Nowadays, due to the improvement of computational systems, 3D geometry of real scenes or objects can find uses for Virtual and Augmented Reality applications even for home applications. KinectFusion \cite{izadi2011kinectfusion} is a system that allows 3D reconstructions through a moving Kinect device and in addition it can be used for augmenting real scene based on its reconstructed geometry.

In \cite{jones2010build} the authors propose a system which enables users to interact with surface particles in real world with the use of a camera, a projector and an infrared pen. The proposed system employs SLS 3D scanning technology in order to reconstruct the 3D geometry of the scene in order appropriately augment the surface particles with the projector. In \cite{bruno20103d} the use of digitization of archaeological findings is discussed for the development of integrated virtual exhibitions.

\section{System Overview}
\label{sec:system_overview}
Figure \ref{fig:system_overview} shows the system overview. Firstly, given as input the resolution of the projector, a sequence of patterns is encoded. Two of the most effective ways of encoding this type of information are presented in Section \ref{sec:encoding_of_patterns}. Secondly, the cameras-projector system is geometrically calibrated and the intrinsic and extrinsic parameters are calculated as described in Section \ref{sec:calib}. Thirdly, the sequence of patterns is projected on the object and images are captured by the cameras for each pattern in the sequence, as described in Section \ref{sec:acquisition}. Finally, the captured images are decoded in order to derive a pixel-to-pixel mapping between the images captured by the cameras as explained in Section \ref{sec:decoding} and is then used to reconstruct a pointcloud of the object using triangulation as described in Section \ref{sec:reconstruction}.

\section{Encoding in Temporal Domain}
\label{sec:encoding_of_patterns}
The first step of the process is the encoding of the information into patterns in the temporal domain. This involves the generation of encoded patterns which when projected in sequence on the object being scanned they allow the unique identification of each of the projector's pixels. In order to achieve this, the encoding of the information is performed in the temporal domain i.e. the encoding is a function of time and all encoded patterns in the sequence are required in order to uniquely identify each pixel.

There are several schemes for encoding information into sequences of patterns, the most popular being the Binary-code and Gray-code encoding which are performed in the temporal domain. In both cases, the information about the two image axes X, Y is encoded separately into a different pattern. A projector $P$ with resolution $P_{res_x}, P_{res_y}$, will result in $N_{cols} = \lceil \log_2 (P_{res_x})  \rceil$ encoded patterns representing the columns, and in $N_{rows} = \lceil \log_2 (P_{res_y}) \rceil$ encoded patterns representing the rows. For example a projector with resolution $1024x768$ will result in $N_{cols} = 10$ and $N_{rows} = 10$ i.e. a total of $20$ patterns.

\subsection{Binary Encoding}
In the binary encoding, the decimal number $D_{c_{i}}$ corresponding to each column $c_{i}$ where the index $i \in P_{res_{x}}$ is converted to its binary form $B_{c_{i}}$. Each individual bit $b_{c_{i}}^{k}$ where $k \in [1, |B_{c_{i}}|]$, of the binary form $B_{c_{i}}$ is then marked in the $k^{th}$ pattern in the sequence, with black color if its value is 0 or white color if its value is 1. Note, that the length of the binary form $B_{c_{i}}$ equals the number of patterns $N_{cols}$. Similarly, this process is repeated for each row $r_{j}$ where $j \in P_{res_{y}}$. 

\begin{figure}[!ht]
 \centering
 \includegraphics[scale=1.6]{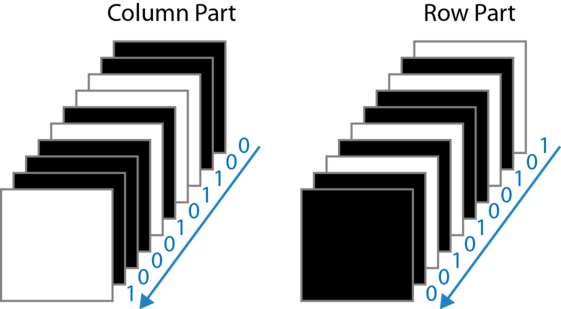}
 \caption{\label{fig:pixel_encoding} Encoding of a single pixel $p$ at location $(209,660)$ is encoded into two sequences of $1x1$ patterns in temporal domain; one sequence representing the column and the other the row information.}
\end{figure}

An example of the process is shown in Figure \ref{fig:pixel_encoding} where the binary form of a single pixel $p$ at location $(209,660)$ is encoded into two sequences of $1x1$ patterns in temporal domain; one sequence representing the column and the other the row information. The column sequence represents the binary form of the column's decimal number i.e. $B_{c_{p}} = 1000101100$ and the row sequence represents the binary form of the row's decimal number i.e. $B_{r_{p}} = 0010100101$. As it can be seen, each pattern corresponds to a bit in the binary pattern and is represented with either the color black or white.

\begin{figure}[!ht]
 \centering
 \includegraphics[scale=1]{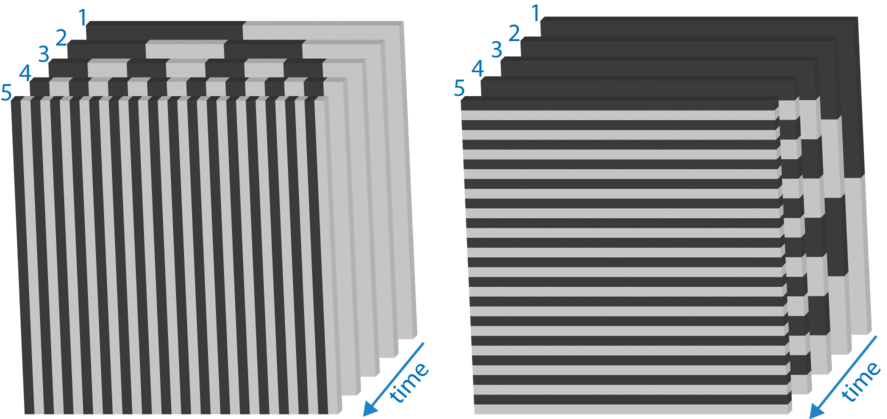}
 \caption{\label{fig:pattern_encoding} Encoding the information into a sequence of patterns in temporal domain.}
\end{figure}

By iteratively performing the process for every pixel $p_{(x,y)}$ in the projector where $x \in res_{x}, y \in res_{y}$, two sequences of 2D patterns are produced representing all the columns and all the rows respectively. Figure \ref{fig:pattern_encoding} shows an example of the final sequences for the columns and rows. The patterns are shown in the order they are projected as indicated by the time arrow i.e the first pattern to be projected corresponds to the most significant bit and appears as the last one in the diagram, followed by the remaining. As time progresses the frequency of the pattern increases until it reaches the highest frequency which represents the least significant bit in binary pattern.

The result is a sequence of binary encoded patterns where each pattern represents one bit of the binary sequence. This produces images containing vertical stripes in the case of the columns, and horizontal stripes in the case of the rows. 

\subsection{Gray-code Encoding}
\begin{figure}[!ht]
	\centering
	\subfloat[\label{fig:bin}]{\includegraphics[scale=0.8]{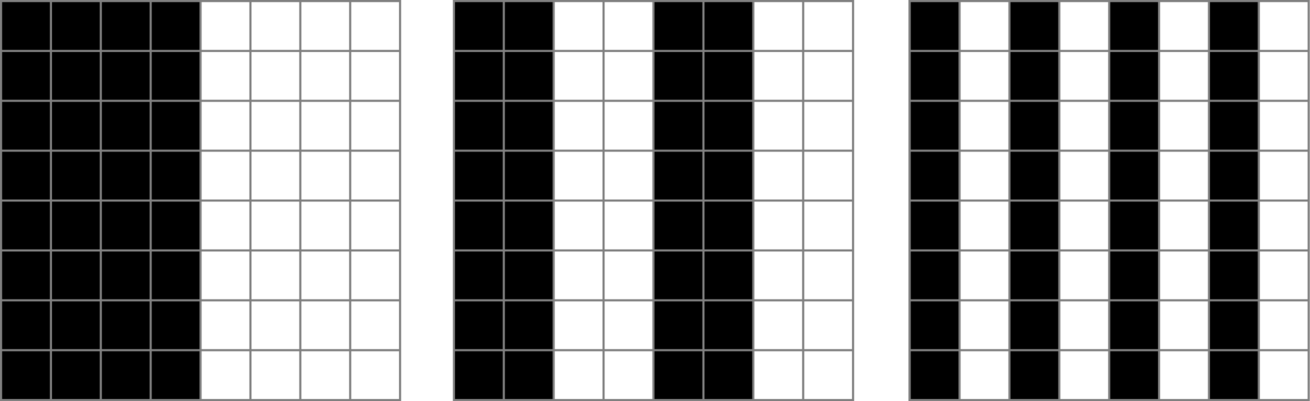}}
	\\
	\subfloat[\label{fig:gray}]{\includegraphics[scale=0.8]{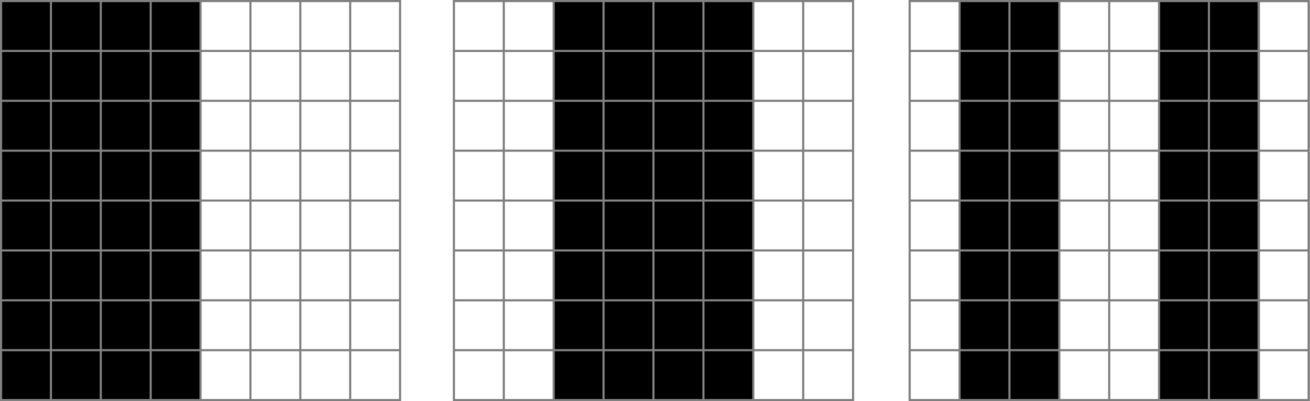}}
	\caption{(a) Binary encoding of the columns for an area of 8x8 pixels. (b) Gray encoding corresponding to the same columns of the area in (a).}
\end{figure}

The Gray-code encoding works in a similar way as the binary encoding previously described, however it ensures that there is only one bit different between consecutive patterns. Figure \ref{tab:gray_code} shows a comparison between Gray and Binary encoding for the columns of an 8x8 area. As it is evident, in the case of Gray codes there is a maximum of one bit difference between any two consecutive encoding of values. Figure \ref{fig:gray} shows a sequence of encoded patterns using Gray-codes, corresponding to the same area of 8x8 as in Figure \ref{fig:bin}.

\begin{figure}[!ht]
\centering
\begin{tabular}[b]{ | c | c | c |}
				   \hline 
				    Decimal Value & Gray-code & Binary code\\
				    \hline                       
 0 &  000  &  000\\
 \hline
 1 &  001  &  001\\
  \hline
 2 &  011  &  010\\
  \hline
 3 &  010  &  011\\
  \hline
 4 &  110  &  100\\
  \hline
 5 &  111  &  101\\
  \hline
 6 &  101  &  110\\
  \hline
 7 &  100  &  111\\
				   \hline  
				 \end{tabular}
				 \caption{\label{tab:gray_code} Comparison between Gray codes and binary codes. In the case of Gray codes there is only one bit difference between any two consecutive encodings.}
\end{figure}

\begin{figure*}[!ht]
	\centering
	\subfloat[\label{fig:bin2gray}]{\includegraphics[scale=1.5]{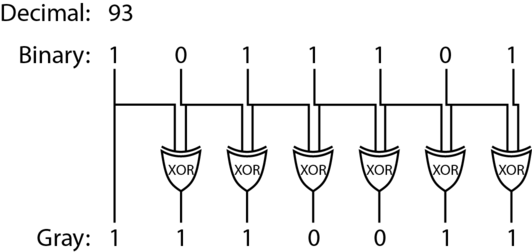}}
	\subfloat[\label{fig:gray2bin}]{	\includegraphics[scale=1.5]{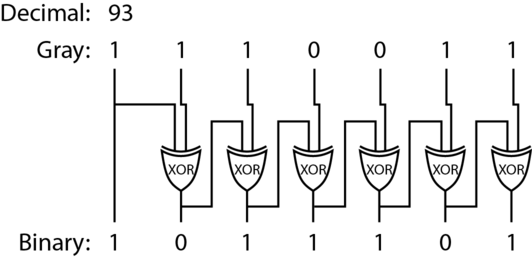}}
	\caption{(a) Encoding of decimal number 93. The Binary Code representing the decimal number 93 is converted into Gray Code. (b) Conversion of Gray Code to Binary Code.}
\end{figure*}

Gray codes can be calculated by first computing the Binary representation of a number and then converting it as follows: copy the most significant bit as is, and then for the remaining bits (taking one bit at a time), replace with the result of an XOR operation of the current bit with the previous bit of higher significance in the binary form. An example is shown in Figure \ref{fig:bin2gray} for the computation of the Gray code for the decimal number 93. 

Similarly, the conversion of a Gray code to the corresponding Binary code is as follows: 
copy the most significant bit as is and for the remaining bits (taking one bit at a time), replace with the result of the XOR between the current bit in the Gray code and the previous bit of higher significance in the Binary code. Figure \ref{fig:gray2bin} shows the computation of the Binary code from the Gray code representing the decimal number 93.

Although the two schemes presented above result in similar patterns, the Gray encoding is considered as a better alternative to the Binary Codes. When using Gray codes the value of the least significant bit changes after two consecutive stripes, in contrast to the Binary codes where the value changes at every pattern. This can be seen in Figure \ref{fig:bin}, \ref{fig:gray} where the pattern using Gray encoding corresponding to the least significant bit, contains stripes with thickness of two pixels, as opposed to the pattern using Binary encoding corresponding to the least significant bit which contains stripes with thickness of a single pixel. Larger stripe thickness is preferable since it can considerably reduce unwanted effects such as color bleeding in the projection or/and in the captured images.

\subsection{Implementation Issues}
The generation of the two sequences of pattern is performed using Gray encoding as described above. However, there are two important implementation issues to consider:
\begin{itemize}
	\item Firstly, the choice of the colors to be used in the patterns. The patterns can have any two colors, although traditionally black and white has been used. In any case, each pixel must have an intensity value which will be used as a threshold to determine the value of the pixel i.e. 1 or 0. Therefore, it is imperative to take this into account when choosing the two colors and choose colors which will result in large intensity differences. A poor choice on colors, can otherwise jeopardize the entire process by failing to distinguish a pixel's value later on in the process. 
	For this reason and to overcome this limitation, it is preferable to project a pattern followed by a projection of its color-inverted pattern. Inverted pattern images are images with the same structure as the original but with inverted colors. This provides an effective method for easily determining the intensity value of each pixel when it is lit (highest value) and when it is not lit (lowest value). The threshold for the intensity value of a pixel $p$ with a highest value $p_{h}$ and a lowest value $p_{l}$ can then be computed as the average $\tau_{p} = (p_{h} - p_{l})/2$. 
	
	\item Secondly, identifying shadow regions. The cameras observe the object from different angles than the projector, hence quite often there will be a viewing areas which lies in shadow regions. Thus, it is preferable that pixels falling under a shadow region are removed at the early stages of the process. This can be achieved by projecting a white and then black image on the object and capturing the images. By evaluating the intensity value of the pixels in the images, one can determine which pixels fall under a shadow region by identifying cases where the intensity values in the two captured images are very similar. A "shadow mask" can then be created which leaves only the pixels which do not fall under shadow regions. This can considerably reduce computational processing time. Section \ref{subsec:shadow_mask} explains how to compute the shadow mask.
\end{itemize}

Thus, the final projection sequence contains the column and row pattern sequences encoded using Gray-code, their inverted patterns, as well as two images of solid colors, one for each used color. The patterns are projected in sequence as follows: first the two solid color images, then interchangeably the column and its inverted sequence, followed by interchangeably the row and its inverted sequence.

\section{Camera Calibration}
\label{sec:calib}

\begin{figure}[!ht]
	\centering
	\includegraphics[scale=1.6]{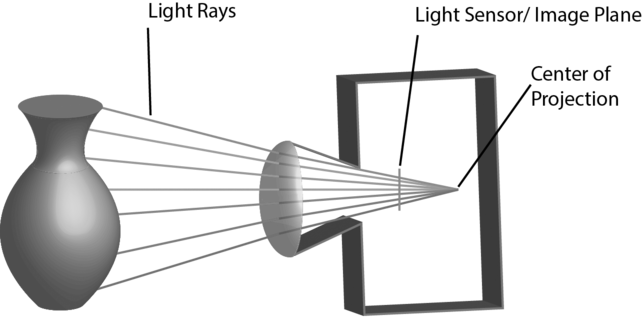}
	\caption{\label{fig:camRays} Image formation. Light rays being reflected in the scene, pass through the camera lens, then through the aperture and interact with the photo-sensitive light sensor.}
\end{figure}

When considering a camera, the image is formed when light rays being reflected in the scene, pass through the camera lens, through the aperture and interact with the photo-sensitive light sensor. If there had been no light sensor present, then the light rays would converge at a single point called the center of projection as shown in Figure \ref{fig:camRays}. A projector can be thought of as an inverted camera. The light rays start from a point i.e. light bulb inside the projector, then go through mirrors and/or lenses and finally interact with the scene.

In both cases (camera, projector), one can relate a light ray to a particular pixel if the geometry of the system at the time the image was taken is known. The geometry is defined in terms of a set of intrinsic and extrinsic parameters which can be computed by performing a geometric calibration.

To calibrate the cameras, an object with known geometric characteristics is used; usually a flat board containing a printed checker pattern. By taking images of the board at different positions and orientations as explained in \cite{zhang2000flexible} we can compute the intrinsic and extrinsic parameters which specify the camera matrix C in Equation \ref{eq:camera_matrix},
\begin{equation}
	C =	
	\underbrace{
		\left[
			\begin{array}{ccc}
				\alpha & -\alpha cot(\theta) & u_{0}\\
				0 & \frac{\beta}{sin(\theta)} & v_{0}\\
				0 & 0 & 1
			\end{array}
		\right]
	}_{intrinsic}
	\underbrace{
		\left[
			\begin{array}{cccc}
				r_{11} & r_{12} & r_{13} & t_{x}\\
				r_{21} & r_{22} & r_{23} & t_{y}\\
				r_{31} & r_{32} & r_{33} & t_{z}
			\end{array}
		\right]
	}_{extrinsic}
	\label{eq:camera_matrix}
\end{equation}

where $\alpha=kf_{x}$, $\beta=kf_{y}$, $(f_{x},f_{y})$ is the focal length on the x and y axis respectively, $\theta$ is the skew angle, $u_{0},v_{0}$ is the principal point on the x and y axis respectively and $r_{1-3}, t_{x-z}$ determine the camera's rotation and translation relative to the world. 

Given a minimum of four 2D to 3D correspondences specified interactively by the operator the camera extrinsic and intrinsic parameters can be accurately estimated. The camera pose estimation is performed using a non-linear Levenberg-Marquardt optimization \cite{opt_1,opt_2} which minimizes the error function $E_{C_{k}}$ for each camera $C_{k}$,
\begin{equation}
	E_{C_{k}} = \frac{1}{n} \sum_{i=0}^n \sqrt{(I_{x}^{i} - P_{x}^{i})^2 + (I_{y}^{i} - P_{y}^{i})^2)}
	\label{eq:minimization_function}
\end{equation}

where $I^{i}$ is the $i$th image point, $P^{i}$ is the projection of the $i$th 3D world point and $n$ is the number of 2D to 3D correspondences. 

In addition to the camera matrix $C$ as defined in Equation \ref{eq:camera_matrix}, the intrinsic parameters include the distortion coefficients; three coefficients for the radial distortion and two coefficients for the tangential distortion. 

Although, theoretically the camera matrix $C$ can be computed from a single image of the flat board, it is recommended that 10-20 images of the flat board at different orientations and positions are used. 

\begin{figure}[!ht]
	\centering
	\includegraphics[scale=1.2]{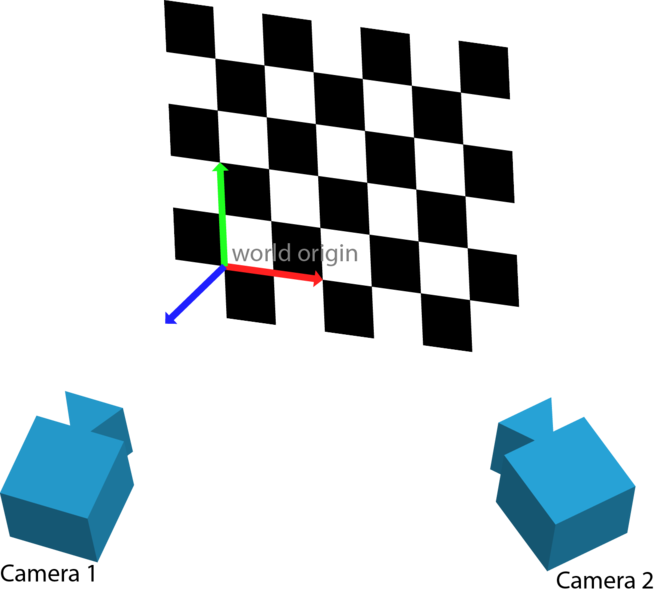}
	\caption{\label{fig:calibration} Multiple Camera Calibration. For each orientation and position of the calibration board, images are captured from all cameras. This allows the use of a common reference point i.e. world origin.}
\end{figure}

\textbf{Multi-Camera System Extrinsics:} In order to calculate the extrinsic parameters of each camera, its intrinsic ones must be already calculated and then using one picture, where the origin of the world is specified, the rotation and translation based on the world origin can be estimated. To achieve this, all cameras in the system must capture the scene at the same time, in order to take the same position and orientation of the board,as shown in Figure \ref{fig:calibration}. In this way the same world origin can be used in order to relate the motion between the cameras i.e. the rotation and translation, as explained later in the Section \ref{sec:reconstruction}.  

\subsection{Implementation Issues}
\begin{itemize}
\item Once the calibration is performed, there should be no movement to any part of the system otherwise a re-calibration will be needed. 
\item The camera calibration in 3DUNDERWORLD-SLS is implemented using OpenCV's calibration functions. Another well-known and established calibration toolbox is the Camera Calibration Toolbox for MatLab \cite{bouguet2004camera}.
\end{itemize}

\section{Acquisition}
\label{sec:acquisition}
The acquisition is a relatively straight forward process. Each image of the sequence is projected on the object and every camera in the system captures an image. However, there are several things to consider:
\begin{itemize}
	\item The object should remain static during the acquisition.
	\item The cameras' and the projector's settings (exposure, brightness etc.) should be adjusted according to the lighting in scene. Indirect light reaching the scene from other sources should be eliminated if possible in order to achieve better results. 
	\item If the acquisition process is automated then it is recommended that there is a "forced-delay" after each image capture until the cameras confirm that the last image was captured and stored successfully. This will reduce and/or eliminate unwanted artifacts which may appear because the projector delayed the projection of a pattern or the camera delayed the capturing an image.
\end{itemize}

\begin{figure}[!ht]
	\centering
	\subfloat[\label{fig:view1}]{\includegraphics[scale=0.20]{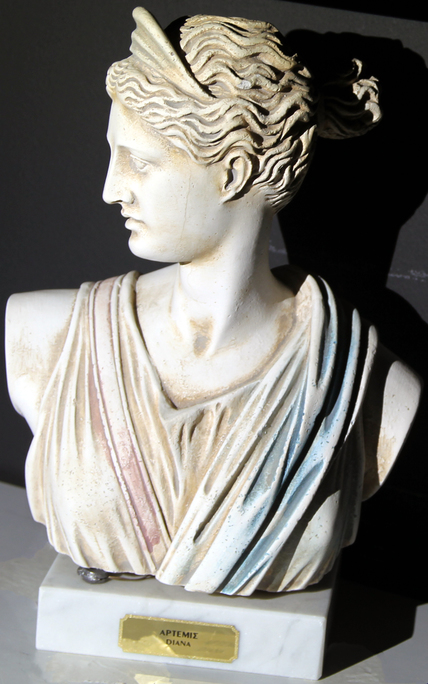}}
	\hfil
	\subfloat[\label{fig:view2}]{\includegraphics[scale=0.20]{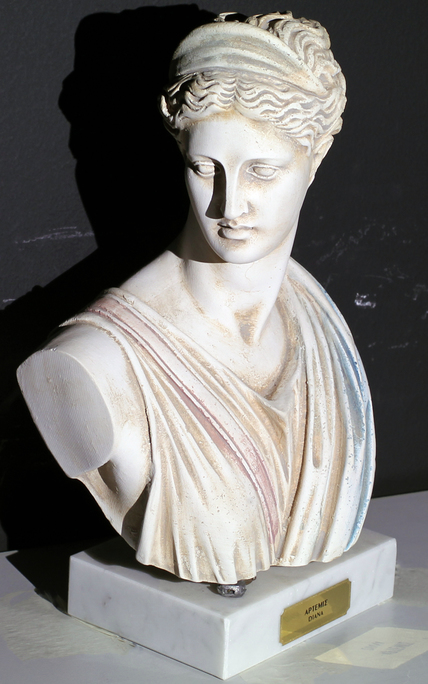}}
	\hfil
	\subfloat[\label{fig:mesh}] {\includegraphics[scale=0.24]{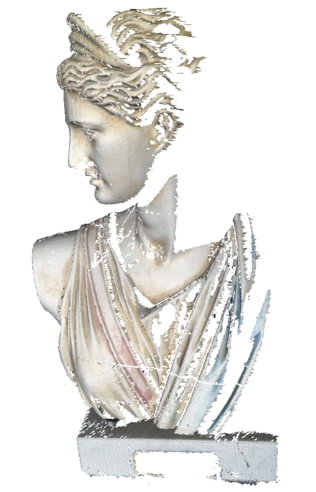}}
	\caption{\label{fig:singleScan}{Result of a single scan. (a) and (b) show the two camera views and (c) the reconstructed mesh. In same cases more scans are needed in order to obtain the desired area.}}
\end{figure}

An acquisition set consists of the images captured by each camera for each pattern in the sequence. However, from each acquisition set, only the parts of the scene which are visible to all the cameras can be reconstructed (Figure \ref{fig:singleScan}), thus more scans may be required to obtain all sides of the object. This can be achieved by changing the object's position and orientation between each acquisition set while ensuring that no changes occur in the cameras and projector.

\section{Decoding of Captured Images}
\label{sec:decoding}

Following the acquisition of the data, the next step is the decoding of the captured images. This involves the calculation of the shadow masks and the decoding of the patterns.
\subsection{Computing the shadow mask}
\label{subsec:shadow_mask}
\begin{figure}[!ht]
	\centering
	\subfloat[]{\includegraphics[scale=0.3]{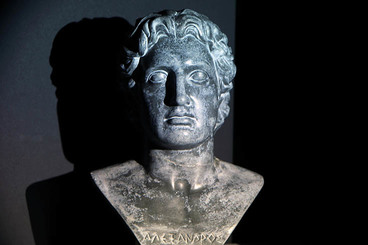}}
	\hfil 
	\subfloat[]{\includegraphics[scale=0.3]{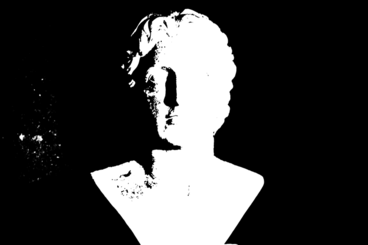}}
	\caption{\label{fig:shadow_mask} Shadow mask example. (a) Image captured by the camera. (b) Shadow mask - all black pixels are removed from subsequent processing.}
\end{figure}

The goal of this process is to determine which pixels in a camera's image fall under shadow regions. This involves comparing each pixel's intensity values between the two first projections: the black and white images. Pixels whose intensity values in the two captured images corresponding to the black image projection and white image projection is large, are considered as a valid pixels. If the difference is small then the pixel is marked as "in-shadow". Figure \ref{fig:shadow_mask} shows an example of a shadow mask.

\subsection{Decoding the patterns}
\label{subsec:decoding_patterns}

\begin{figure}[!ht]
 \centering
 \includegraphics[scale=1.8]{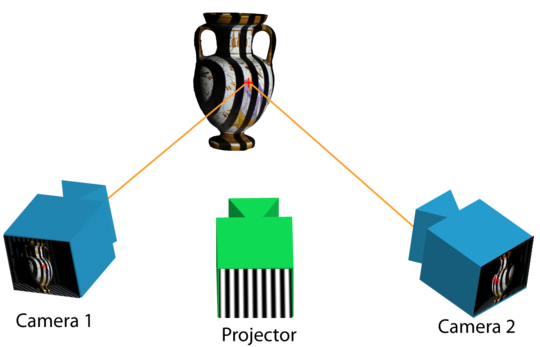}
 \caption{\label{fig:decoding} A 3D point is viewed by two different cameras. The decoding aims at deriving a mapping between the pixels of the two cameras, corresponding to the same 3D point.}
\end{figure}

The cameras capture images of the encoded projected patterns. The next step is to decode each pixel in the captured images into their corresponding decimal number representing the column and the row. This provides a mapping between the pixels in the cameras i.e. pixels in the captured images which correspond to the same projector pixel as shown in Figure \ref{fig:decoding}.

More specifically, the decoding is performed as follows: 
\begin{itemize}
	\item Determine whether a pixel $p$ is lit or not (1 or 0) in the images capturing the projected sequence of patterns encoding the columns.
	\item Calculate its binary form $B_{p}$.
	\item Convert the binary form $B_{p}$ into the equivalent decimal number $x$.
\end{itemize}
Similarly, the process is repeated for the images capturing the projected sequence of patterns encoding the rows and results in a decimal number $y$. Thus, $(x,y)$ are the image coordinates of the projector's pixel corresponding to the pixel being decoded in a camera. By repeating the process for all the cameras, we can map the pixels not only to the projector's pixels but rather to the other cameras viewing the object. It should be noted that a camera pixel may be mapped to more than one pixels of another camera due to differences between the camera and projector resolutions.

\section{Reconstruction}
\label{sec:reconstruction}

The decoded captured images result in a set of a many-to-many mappings between the pixels of the different cameras. Next, by triangulating the rays corresponding to each pair, a 3D point is computed at their intersection. The projection of this point falls onto the mapped pixels in the different cameras.

\subsection{Pixel-to-Ray}
\label{subsec:pixel_to_ray}
A camera ray is a straight line in 3D Euclidean space that starts from the camera's center of projection (Figure \ref{fig:camRays}), intersects the image plane and extends outward to the scene. The ray can be defined with a single point of the ray and the ray's direction vector. 

In order to compute the direction vector two ray's points are needed. The first point is the camera's center of projection $q_{center}^{camera} = (0,0,0)$. The second point is the point corresponding to the pixel from which the ray passes through. That point can be computed by first finding the undistorted pixel's position, using the distortion coefficients computed during camera calibration, and then "unprojecting" the pixel into 3D space. This is achieved by multiplying the inverse of the camera matrix $C$ with the pixel's coordinates as given by the Equation \ref{eq:pixel_unproject},
\begin{equation}
q_{pixel}^{camera} = C^{-1} \times \begin{pmatrix}
x \\ 
y \\ 
1
\end{pmatrix}
\label{eq:pixel_unproject}
\end{equation}

Therefore, both $q_{center}^{camera}$ and $q_{pixel}^{ccamera}$ are in the camera's local coordinates frame and they must be converted to world coordinates using the extrinsic parameters computed during the camera calibration as follows,
\begin{equation}
	\begin{array}{lcl}
		q^{world} & = & R^{-1} \times q^{camera} - R^{-1} \times T
	\end{array}
	\label{eq:cam2world}
\end{equation}
where $R$ is the rotation matrix and $T$ is the translation vector of the camera relative to the world origin shown in Figure \ref{fig:calibration}.

Thus, the ray corresponding to pixel $p$ is defined as $R_{p}^{world} = < p_{center}^{world}, \vec{v}_{dir}^{world}>$ in world coordinates, where $\vec{v}_{dir}^{world} = p_{pixel}^{world} - p_{center}^{world}$.

\subsection{Ray Triangulation}
\label{subsec:ray_triangulation}

Given one pixel in image $I_{1}$ and its corresponding pixel in image $I_{2}$, two rays are formed as described above, and their intersection is computed. In practice, the triangulation of two rays in 3D space can be considered as an ill-posed problem since quite often the two rays do not intersect 'exactly' but rather pass by one another in close proximity. 

\begin{figure}[!ht]
	\centering
	\includegraphics[scale=2]{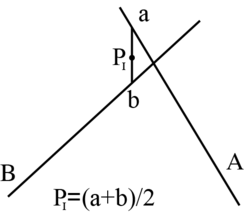}
	\caption{\label{fig:raysNotInters} Ray intersection. In cases where two rays do not intersect 'exactly', the intersection point is considered to be the midpoint of the shortest segment connecting the two rays.}
\end{figure}

In order to overcome this limitation, the segment perpendicular to the two rays with the shorted distance is computed, and the middle point of the segment is considered to be the intersection point. Figure \ref{fig:raysNotInters} shows an example where the segment $ab$ is the shortest line connecting the two rays $A$ and $B$. The mid-point $p$ is considered to be the intersection point.

\textbf{Computation of the Intersection Point:} Consider two lines in 3D space $A$ and $B$ passing through points $p$ and $q$, with direction vectors $\vec{u}$ and $\vec{v}$ respectively, and let the two closest points on the lines be $a$ and $b$, as defined in Equation \ref{eq:line1} and Equation \ref{eq:line2}, where $s$ and $t$ are scalar values.
\begin{equation}
	a = p + s*\vec{u}
	\label{eq:line1}
\end{equation} 

\begin{equation}
	b = q + t*\vec{v}
	\label{eq:line2}
\end{equation}
The segment connecting $a$ and $b$ is perpendicular to the lines, hence the dot product of their vectors is equal to $0$ as follows,  
\begin{equation}
	(a-b).\vec{u}=0
	\label{eq:dot1}
\end{equation} 
\begin{equation}
	(a-b).\vec{v}=0
	\label{eq:dot2}
\end{equation}
and with Equations \ref{eq:line1} and \ref{eq:line2}, the Equations \ref{eq:dot1} and \ref{eq:dot2} become,
\begin{equation}
	\vec{w}.\vec{u} + s*\vec{u}.\vec{u} - t*\vec{v}.\vec{u} = 0
	\label{eq:exp1}
\end{equation} 
\begin{equation}
	\vec{w}.\vec{v} + s*\vec{v}.\vec{u} - t*\vec{v}.\vec{v} = 0
	\label{eq:exp2}
\end{equation} 
where
\begin{equation}
	\vec{w} = p - q
	\label{eq:w}
\end{equation} 
From Equation \ref{eq:exp1} and Equation \ref{eq:exp2} it follows that, 
\begin{equation}
	s = \frac{\vec{w}.\vec{u}*\vec{v}.\vec{v} - \vec{v}.\vec{u}*\vec{w}.\vec{v}}{\vec{v}.\vec{u}*\vec{v}.\vec{u} - \vec{v}.\vec{v}*\vec{u}.\vec{u}} 
	\label{eq:seq}
\end{equation}
\begin{equation}
	t = \frac{\vec{v}.\vec{u}*\vec{w}.\vec{u} - \vec{u}.\vec{u}*\vec{w}.\vec{v}}{\vec{v}.\vec{u}*\vec{v}.\vec{u} - \vec{v}.\vec{v}*\vec{u}.\vec{u}} 
	\label{eq:teq}
\end{equation}
The intersection point is the average of the 2 end-points of the segment as follows,
\begin{equation}
	P_i = \frac{(p + s*\vec{u}) + (q + t*\vec{v})}{2} 
	\label{eq:final}
\end{equation}
Thus, for all ray-pairs the intersections are computed as above in order to avoid any possible problems with non-intersecting rays. 

\subsection{Implementation Issues}
\label{subsec:discussion}
There are several limitations with the chosen triangulation method that should be taken into account. Firstly, although two rays (as defined in our system) should not be parallel, it is a good practice to check for near-parallel conditions as well. In order to to so, it is suggested that a near zero condition is checked on the denominator of the Equations \ref{eq:seq} and \ref{eq:teq}.
 
Another important note is the fact that in practice the mapping between pixels is not one-to-one but rather many-to-many i.e. an area of pixels in one image maps onto another area of pixels in another image. This is due to the fact that the cameras and projector resolutions are different. In situations like these it is suggested that the average of the intersection points is taken as the final intersection point. Averaging the resulting intersection points not only reduces the amount of generated geometry by removing duplicates bu also results in smoother geometry.

Finally, the linear triangulation described in \ref{subsec:ray_triangulation} is the simplest triangulation method and not necessarily the best one. In \cite{hartley1997triangulation} the triangulation problem is discussed extensively, and the authors propose an alternative method which outperforms the linear triangulation discussed above. We have found through extensive testing that linear triangulation performs very well; at least in our case.

\section{Point Cloud to Mesh}
\label{sec:point2Mesh}
Thus far, we have described how to generate a cloud of 3D points in Euclidean space that represents the scanned scene. In many cases this type of output is not easily usable, since it is difficult to manipulate. For this reason, we may first need to convert it to a 3D mesh, where the points (vertices) are interconnected (with edges) to each other producing faces. Many methods already exist with various complexities and resulting output qualities. In this paper a quick conversion method will be described which takes advantage of the spatial relation between the 3D points and the projector's pixels.

\begin{figure}[!ht]
	\centering
	\includegraphics[scale=1.2]{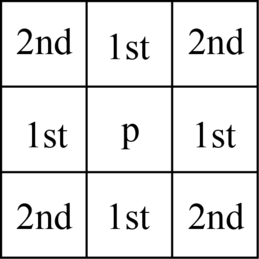}
	\caption{\label{fig:neighboringPixels} Neighbouring pixels. 1st level neighbours of a pixel $p$ are the previous, next, above and below pixels. 2nd level are the diagonal pixels.}
\end{figure}

As already mentioned, each point in the cloud corresponds to a projector's pixel; in fact each point is representing the area in the scene that was lit by the related projectors' pixel. Based on this fact, areas that are lit by neighbouring pixels are next to each other, and similarly the points relating to those pixels are neighbours as well. Neighbours of a pixel $p$ are considered the 8 surrounding pixels, where the ones placed in the same row or column with $p$ are 1st level neighbours, while the diagonals are 2nd level neighbours as it can be seen in Figure \ref{fig:neighboringPixels}.

\begin{figure}[!ht]
	\centering
	\includegraphics[scale=1.6]{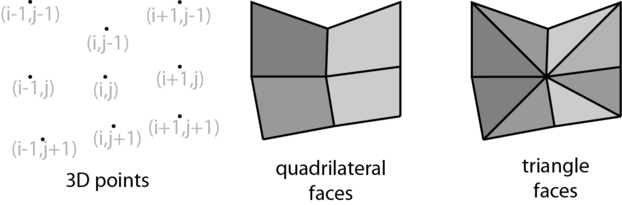}
	\caption{\label{fig:mesh49} Conversion of point-cloud to mesh. 1st level connections result to a quadrilateral faces. 1st and 2nd level result to triangular faces }
\end{figure}

Points of neighbouring pixels are iteratively connected to each other forming faces. Most commonly, faces can be either triangular or quadrilateral. In order to form triangular faces both 1st and 2nd level neighbours are interconnected, in contrast to quadrilateral forming where only 1st level neighbour connections are needed as shown in Figure \ref{fig:mesh49}. Meshes consisting of triangular faces have higher complexity than the quadrilateral ones, however they often appear to form smoother surfaces. 

The described method can introduce some artifacts in the output mesh in cases where some pixels are not visible by more than one camera in which case holes may appear in the mesh, or cases where neighbouring pixels are projected on different surfaces in the scene resulting connections between them. However, this method is generic and quite fast since it is performed in projector's image space and works very well for most of the scene's area. Possible hole problems can be filled by another scanning and possible connections between wrong faces which are seldom found can be easily removed manually. In addition, it can accommodate extra information such as the color of the mesh and it is even suitable for cases where the scanned object is decorated with structured color motives or other images.

\section{Experimental Results}
\label{sec:results}

\subsection{Apparatus}
\label{subsec:apparatus}
In this work we focus on a system consisting of two digital cameras with manual (fixed) focus mode (as opposed to autofocus) and a DLP projector. \footnote{3DUNDERWORLD-SLS v1.0 requires one web camera. \\ 3DUNDERWORLD-SLS v2.x requires one Canon SLR camera.\\3DUNDERWORLD-SLS v3.x requires two or more Canon SLR cameras.\\3DUNDERWORLD-SLS v4.x requires two or more cameras and includes a CUDA GPU implementation as well as a CPU implementation in case an Nvidia card is not found. In this version, we provide a generic camera interface implementation and which the programmer can extend to support any kind of camera.} It is important that the projector's pixels are clearly captured in the images taken by the cameras, thus it is preferable to use a camera which has higher resolution than the resolution of the projection. This however will not affect the resolution of the final 3D model since the reconstruction takes place in projector's space i.e. it will have the resolution of the projector. The process of capturing the images can be done either automatically using a connected computer, or manually. In the latter case, it is imperative that the setup remains static. Even a slight movement can result in misalignment and severe reconstruction errors.

In the open-source scanning system 3DUNDERWORLD-SLS v3.x, we employ two Canon SLR cameras EOS-1D Mark IV with a resolution of 4896x3264, and an inFocus IN110 portable projector with a resolution of 1024x768. All three devices are connected to a portable computer which runs the software for the scanning process. Each of the encoded images (total of 42 for a resolution of 1024x768), generated as previously explained, are projected on the object and at the same time each connected camera captures a photo of the scene. At the end of the scanning process these captured images will be processed in order to compute the 3D geometry of the scene.

\begin{figure}[!ht]
	\centering	
	\includegraphics[scale=0.48]{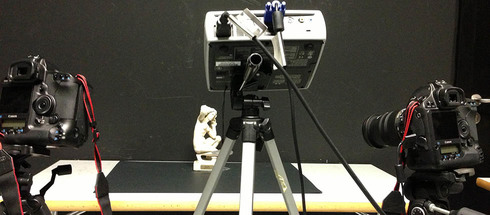}
	\caption{\label{fig:setup} Example of two cameras setup, with the cameras left and right of the projector.}
\end{figure}

\vspace{-20pt}
\subsection{Setup}
\label{subsec:setup}
As this technique is very general, there are several acceptable ways to position the imaging devices. The positioning is subjected to the size of the scanned object, the number of the cameras used and the type of their lenses. Generally it is good practice to have the cameras in non-parallel positions in order to get smoother results. Note however that as the angle between the cameras increases the areas of the object that can be reconstructed decreases since the area visible to \textbf{all} cameras reduces. For this reason, we suggest to position the projector in the middle of the cameras as shown in Figure \ref{fig:setup}.

Another important aspect is to adjust the optics to the object's size. For better results, the projector can be set in a way such that most of the area being lit lies on the object; the more 'lit' pixels project on the object the higher the resolution of the resulting pointcloud. The same rule of thumb applies to the cameras too: it is crucial to ensure that cameras are in close proximity such that the projection is visible down to the pixel level. Additionally, the more camera pixels capture one projector's pixel the smoother the reconstructed model will be. 

After the placement of the imaging devices, it is important to set the focus on the object's surface in order to have a sharp projection and sharp captured images. Before starting the scanning process the system must be calibrated. 

\subsection{Scanning \& Reconstruction}
\label{subsec:scanning_reconstruction}

The proposed algorithms and implemented techniques were extensively tested and the results are reported. All results shown below were generated with the developed open-source scanning system 3DUNDERWORLD-SLS.  Several objects were scanned and information about their structural size in real-life, the number of points and the number of triangles of the reconstructed models are presented in Table \ref{tab:metric1} and Table \ref{tab:metric2}.

Figure \ref{fig:amphora} shows a render of a genuine replica of an amphora (right) next to a photo of the original (left). The amphora has dimensions $18cm x 10cm x 10cm$ and the resulting object contains $1276766$ points and $2553536$ faces.

\begin{table}
\centering
\hspace*{-2pt}\makebox[\linewidth][c]{
\begin{tabular}{| c |  m{0.8cm} | m{0.8cm} | c | c | c |}
	\hline
	\textbf{Model} & \textbf{Real} & \textbf{Recon.} & \textbf{Size} & \textbf{Pts} & \textbf{Tri.} \\
	\hline
	Aphrodite & 
	\includegraphics[scale=0.1]{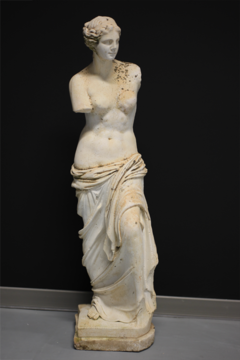} & 
	\includegraphics[scale=0.1]{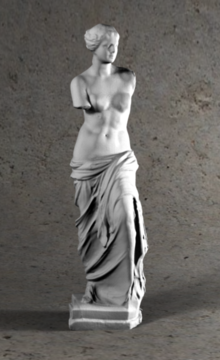} & 
	\begin{tabular}{c}
		$88cm \times$ \\ $20cm \times$ \\ $20cm$	
	\end{tabular} & 
	$602685$ & 
	$1205406$ \\
	\hline
	Aztec & 
	\includegraphics[scale=0.1]{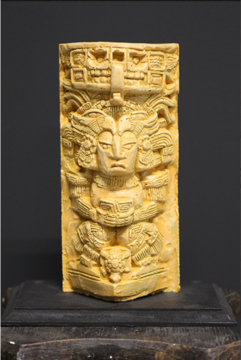} & 
	\includegraphics[scale=0.1]{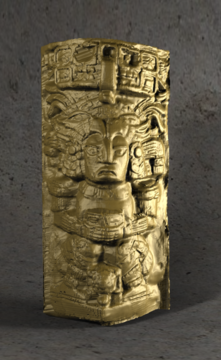} & 
	\begin{tabular}{c}
		$22.5cm \times$ \\ $10.5cm \times$ \\ $6cm$	
	\end{tabular} & 
	$540745$ & 
	$1054972$ \\
	\hline
	Caryatid & 
	\includegraphics[scale=0.1]{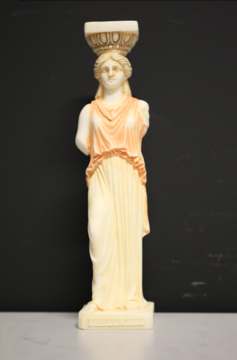} & 
	\includegraphics[scale=0.1]{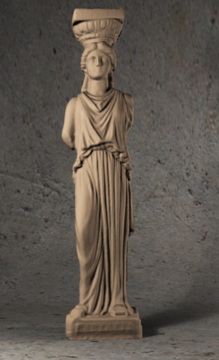} & 
	\begin{tabular}{c}
		$24.5cm \times$ \\ $5.5cm \times$ \\ $5.5cm$	
	\end{tabular} & 
	$659777$ & 
	$1319544$ \\
	\hline
	Aphrodite & 
	\includegraphics[scale=0.1]{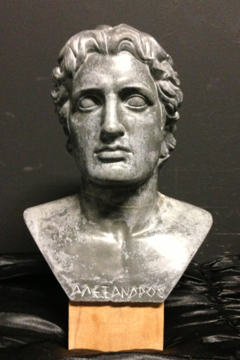} & 
	\includegraphics[scale=0.1]{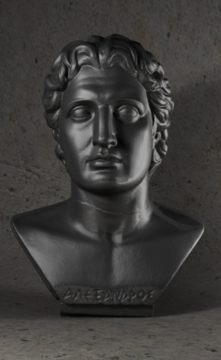} & 
	\begin{tabular}{c}
		$25.5cm \times$ \\ $19cm \times$ \\ $14.5cm$	
	\end{tabular} & 
	$2028162$ & 
	$4056320$ \\
	\hline
\end{tabular}
}
\caption{Videos of the reconstructed models can be found at: \url{http://www.vimeo.com/TheICTLab}}
\end{table}

\begin{figure}[!ht]
	\centering
	\subfloat[]{\includegraphics[scale=0.5]{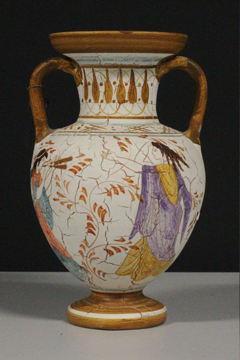}}
	\hfil
	\subfloat[]{\includegraphics[scale=0.5]{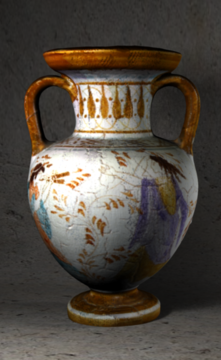}}
	\caption{\label{fig:amphora} Reconstructed model of a genuine replica of an amphora. (a) Photo of the object. (b) A render of the reconstructed model.}
\end{figure}

\section{Evaluation}
\label{sec:evaluation}
The evaluation of the proposed technique is performed by measuring the following four evaluation metrics: linearity, orthogonality, sampling rate, accuracy.

\subsection{Linearity metric}
\label{subsec:linearity}
A perfectly flat plane $\Pi$ is scanned and a plane $\Pi_{fitted} = \langle \alpha, \beta, \gamma, \delta \rangle$ with normal $N_{\Pi_{fitted}} = \langle n_{x}, n_{y}, n_{z} \rangle$ is fitted on the resulting $\aleph$ 3D points. The plane fitting is perfomed using RANSAC and the average error $E_{avg}$ and average $RMSE$ is computed as follows,
\vspace{-10pt}
\begin{equation}
	E_{avg} = \sum_{i=0}^{\aleph} |\delta - (\alpha \times \aleph_{i}^{x} + \beta \times \aleph_{i}^{y} + \gamma \times \aleph_{i}^{z})|/ ||\aleph||
\end{equation}
\vspace{-10pt}
\begin{equation}
	RMSE = \sqrt[]{\frac{\sum_{i=0}^{\aleph} \lbrace \delta - (\alpha \times \aleph_{i}^{x} + \beta \times \aleph_{i}^{y} + \gamma \times \aleph_{i}^{z}) \rbrace^2}{||\aleph||}}
\end{equation}

We scan several different planar surfaces shown in Table \ref{tab:metric1} and perform plane fitting using RANSAC. This metric is measured as the distance of all points from the fitted surface in terms of the average error $E_{avg}$ and the root-mean-square error $RMSE$. As it can be seen from the reported measured the error is minuscule when considering the total number of points. It should be noted that for the box example, the three orthogonal planes were measured separately.

\begin{table*}[!ht]
\hspace*{-10pt}\makebox[\linewidth][c]{
\begin{tabular}{| l | >{\centering\arraybackslash} m{2.5cm} | >{\centering\arraybackslash} m{2.5cm} | c | c | c |}
   \hline   
   Object & Left Image & Right image & \multicolumn{2}{|c|}{Linearity} & Points \\        
   \hline
   & & & $E_{avg}$ & $RMSE$ & $\aleph$ \\
   \hline
   Flat Plane $\Pi_{1}$ & \includegraphics[scale=0.1]{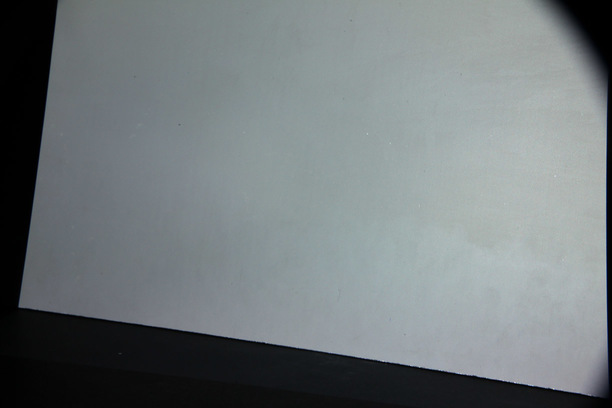} & \includegraphics[scale=0.1]{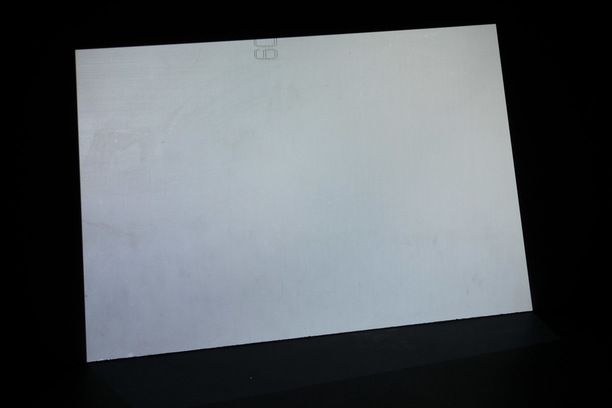} & 0.0046794643 & 0.0083896662 & 490625\\
  \hline
   Flat Plane $\Pi_{2}$ & \includegraphics[scale=0.1]{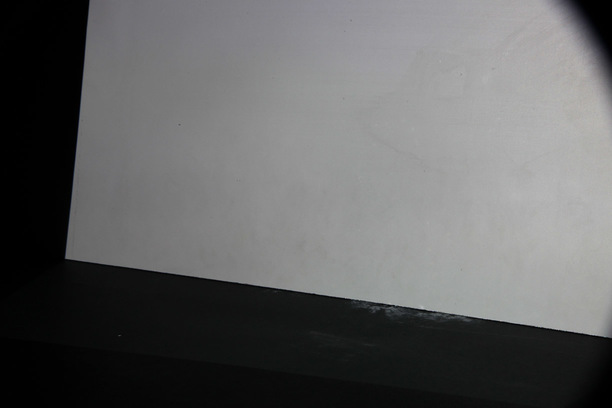} & \includegraphics[scale=0.1]{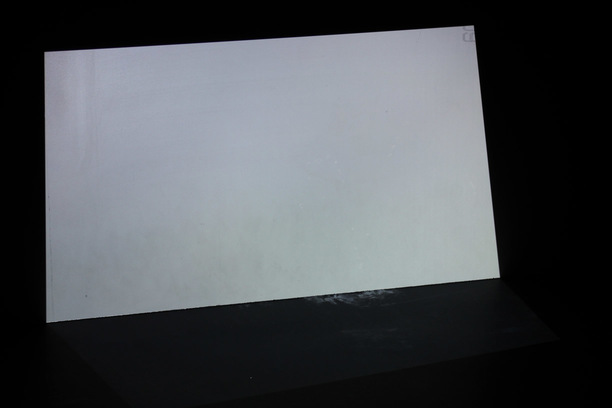} & 0.0109797063 & 0.0186112309 & 359659\\
   \hline
   Flat Plane $\Pi_{3}$ & \includegraphics[scale=0.1]{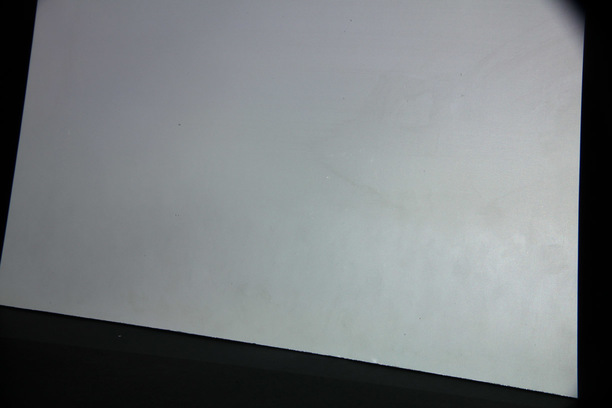} & \includegraphics[scale=0.1]{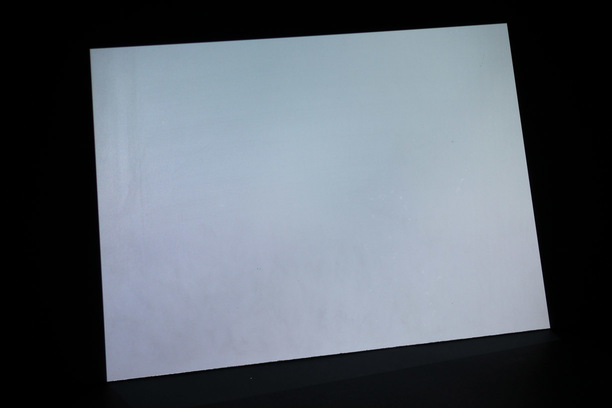} & 0.002567169 & 0.004390225 & 540160\\
   \hline
   Box Plane $\Pi_{Box}^{1}$ & \includegraphics[scale=0.1]{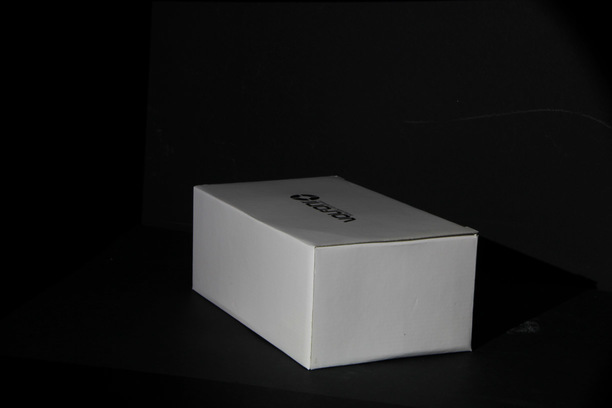} & \includegraphics[scale=0.1]{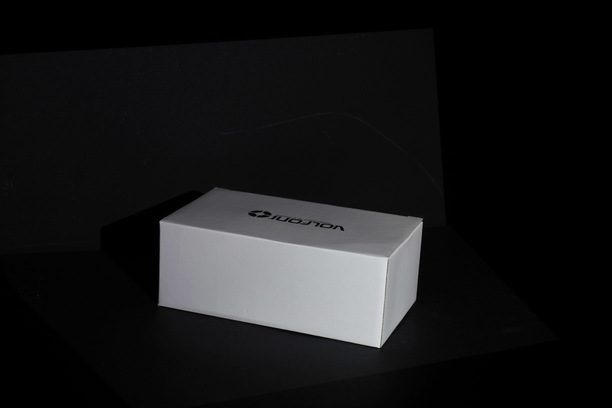} & 0.0012361568 & 0.0015234051 & 59210\\
   \hline
   Box Plane $\Pi_{Box}^{2}$ & \includegraphics[scale=0.1]{./box_cam1.JPG} & \includegraphics[scale=0.1]{./box_cam2.JPG} & 0.0064742412	& 0.0090599699 & 24980\\
   \hline
   Box Plane $\Pi_{Box}^{3}$ & \includegraphics[scale=0.1]{./box_cam1.JPG} & \includegraphics[scale=0.1]{./box_cam2.JPG} & 0.0264965185	& 0.0352513109 & 24464\\
   \hline  
\end{tabular}
}
  \caption{\label{tab:metric1} Linearity metric. This metric is measured by fitting planes using RANSAC to the data and measuring the distance of all points from the fitted surface in terms of the average error $E_{avg}$ and the root-mean-square error $RMSE$.}
  \end{table*}

\subsection{Orthogonality metric}
A set of three perpendicular planes were scanned and three planes $\Pi_{1}, \Pi_{2}, \Pi_{3}$ are fitted respectively using RANSAC. The orthogonality metric is defined as the magnitude of the three dimensional vector containing the measured angles between the three planes in terms of the dot product as follows,
\begin{multline}
	E_{ortho} = \langle dot(N_{\Pi_{fitted}}^1, N_{\Pi_{fitted}}^2), dot(N_{\Pi_{fitted}}^1, N_{\Pi_{fitted}}^3), dot(N_{\Pi_{fitted}}^2, N_{\Pi_{fitted}}^3) \rangle
\end{multline}

We scan the box object containing orthogonal planes shown in Table \ref{tab:metric1}. For each of the three planes a linear surface is fitted using RANSAC. This metric is measured as the angle formed between the three planes as shown in Table \ref{tab:metric2}. As it is evident from the reported results the resulting planes are perpendicular (up to at least the fourth decimal point).

\begin{table*}[!ht]
\centering
\begin{tabular}{| c | c | c | c |}
   \hline   
   $\perp$ & Box Plane $\Pi_{Box}^{1}$ & Box Plane $\Pi_{Box}^{2}$ & Box Plane $\Pi_{Box}^{3}$\\
   \hline
   Box Plane $\Pi_{Box}^{1}$ (long) & - & $89.9997272431^\circ$	& $89.9995961527^\circ$\\
   \hline
   Box Plane $\Pi_{Box}^{2}$ (top) & $89.9997272431^\circ$ & - & $89.9978673427^\circ$\\
   \hline
   Box Plane $\Pi_{Box}^{3}$ (side) & $89.9995961527^\circ$ & $89.9978673427^\circ$ & - \\
   \hline
\end{tabular}
  \caption{\label{tab:metric2} Orthogonality metric. This metric is measured by fitting planes using RANSAC to the three sides of the box shown in Table \ref{tab:metric1} and measuring the angle formed between them.}
\end{table*}

\subsection{Accuracy}
An object containing a ruler of length $L_{orig}$ is scanned. The size of ruler is measured in the reconstructed model as $L_{scan}$. The accuracy is defined as the absolute difference between the two measurements, 
\begin{equation}
	E_{acc} = |L_{orig} - L_{scan}|
\end{equation}
All calibration parameters are calculated in millimeters, hence the measuring units for accuracy is also millimeters.

We scan a planar object which contains a imprinted ruler on its surface. We measure several distances between points on the ruler on the reconstructed model and compute the average accuracy as shown in the Table \ref{tab:metric3}.

\begin{table}[!ht]
\centering
\begin{tabular}{| c | c |}
   \hline
   \multirow{20}{*}{\includegraphics[scale=0.6]{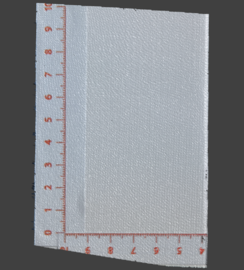}} &    $E_{acc}$ = 0.0251\\
&$E_{acc}$ = 0.0827\\
&$E_{acc}$ = 0.0259\\
&$E_{acc}$ = 0.0166\\
&$E_{acc}$ = 0.0716\\
&$E_{acc}$ = 0.04335\\
&$E_{acc}$ = 0.0533\\
&$E_{acc}$ = 0.1223\\
&$E_{acc}$ = 0.003\\
&$E_{acc}$ = 0.0361\\
&$E_{acc}$ = 0.0413\\
&$E_{acc}$ = 0.1124\\
&$E_{acc}$ = 0.0397\\
&$E_{acc}$ = 0.0174\\
&$E_{acc}$ = 0.1091\\
&$E_{acc}$ = 0.0298\\
&$E_{acc}$ = 0.0933\\
&$E_{acc}$ = 0.0163\\
&$E_{acc}$ = 0.0153\\
&$E_{acc}$ = 0.0855\\
   \hline
   $E_{avg} = \sum\limits_{i=0}^{20} \frac{E_{acc}^{i}}{20}$ & 0.0520025\\
   \hline
\end{tabular}
  \caption{\label{tab:metric3} Accuracy metric. This metric is measured by taking multiple measurements on a reconstructed planar object with an imprinted ruler on its surface.}
  \end{table}

\subsection{Sampling rate}
\begin{figure}[!ht]
	\centering	
	\includegraphics[scale=0.7]{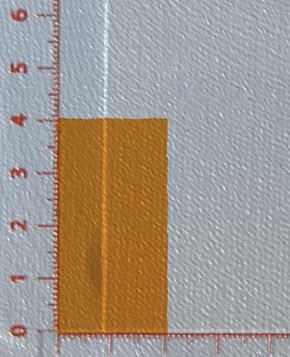}
	\caption{\label{fig:area_selection} Example of a density measurement on the reconstructed object(rendered). Multiple areas are selected on a marked reconstructed object and the number of points and number of faces(triangles) are counted in order to compute the point and face density per squared centimetre. }
\end{figure}

\begin{figure}[!ht]
	\centering
	\subfloat[]{\includegraphics[scale=0.9]{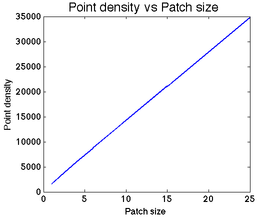}}
	\hfill
	\subfloat[]{\includegraphics[scale=0.9]{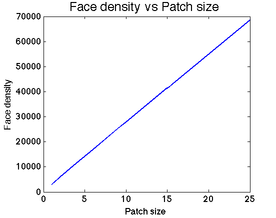}}
	\caption{\label{fig:density} The point and face density for difference patch sizes. \textbf{Mean point density per $cm^2$: 1458.3}. \textbf{Mean face density per $cm^2$: 2835.6}}
\end{figure}

The sampling rate is computed by selecting an area(patch) of known dimensions $(width, height)$ in the reconstructed model and measuring the number of points contained. We scanned a planar object containing an imprinted ruler on its surface. We manually crop multiple small patches from the reconstructed object and measure the corresponding point and face densities, an example of this procedure is shown in Figure \ref{fig:area_selection}. We report the individual point densities per squared centimeter ($cm^2$) and the average point density per squared centimeter ($cm^2$) in Table \ref{tab:metric4}. As it can be also be seen in Figure \ref{fig:density} the mean face density per $cm^2$ is 2835.6 faces and the mean point density per $cm^2$ is 1458.3. It should be noted that all faces are triangles for these experiments.

\begin{table}[!ht]
\centering
\begin{tabular}{| c | c | c | c |}
   \hline
   \textbf{ Width ($cm$)} & \textbf{ Height ($cm$)} & Point density & Face density\\
   \hline
	1 & 1 & 1588 & 2995\\
	\hline
	2 & 2 & 5968 & 11594\\
	\hline
	3 & 5 & 21129 & 41551\\
	\hline
	3 & 3 & 12864 & 25211\\
	\hline
	4 & 4 & 22354 & 44020\\
	\hline
	1 & 2 & 3117 & 5955\\
	\hline
	5 & 3 & 20959 & 41197\\
	\hline
	2 & 1 & 3060 & 5838\\
	\hline
	3 & 2 & 8700 & 16952\\
	\hline
	2 & 3 & 8740 & 17024\\
	\hline
	5 & 5 & 34759 & 68648\\
	\hline
	5 & 2 & 14295 & 27952\\	
   \hline
	2 & 5 & 14308 & 28004\\
   \hline
   \multicolumn{4}{|c|}{\textbf{Mean point density per $cm^2$: 1458.3}}\\
   \hline
   \multicolumn{4}{|c|}{\textbf{Mean face density per $cm^2$: 2835.6}}\\
   \hline
\end{tabular}
  \caption{\label{tab:metric4} Sampling rate. Multiple patches of difference sizes are taken from the reconstructed model shown in Table \ref{tab:metric3} and their corresponding point and face densities are measured.}
  \end{table}
  
\subsection{Comparison to a High-End Commercial 3D Scanner}
\begin{landscape}
\begin{figure}[!ht]
	\centering
	\subfloat[\label{fig:laser_scan}]{\includegraphics[scale=0.5, angle=-90]{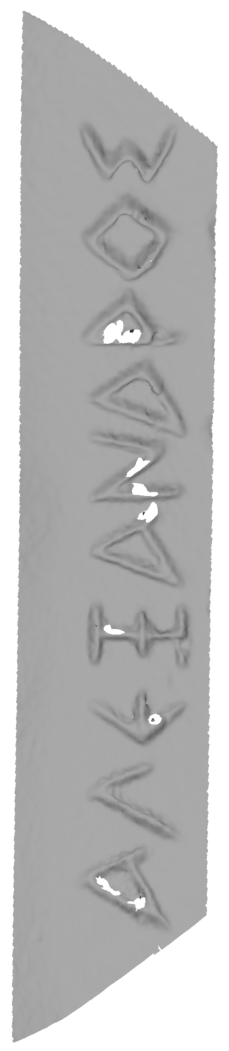}}
	\hfill
	\subfloat[\label{fig:sls_scan}]{\includegraphics[scale=0.5, angle=-90]{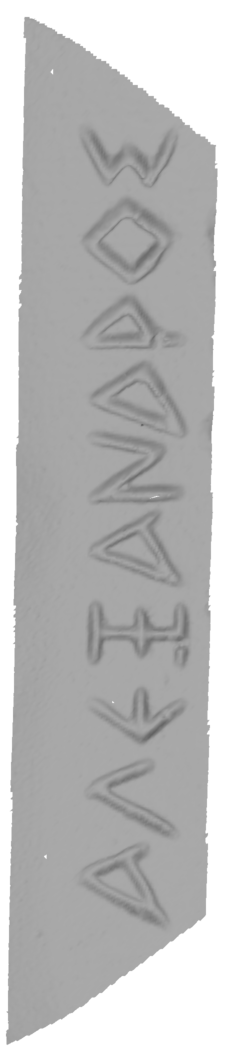}}
	\caption{\label{fig:amphora} The same part of an object reconstructed with (a) ZScanner 700CX configured to the highest resolution and scanned from multiple directions and (b) the proposed system using a \textbf{single scan}. Both meshes were hole-filled using the same settings and are presented with the same Gouraud shading.}
\end{figure}
\end{landscape}

The proposed algorithms and implemented techniques were evaluated against a high-end commercial laser scanner. Figure \ref{fig:laser_scan} shows part of a mesh scanned with a high-end commercial 3D scanner, namely ZScanner 700CX. The laser scanner is hand held and it was used exactly as recommended in its manual with the highest possible resolution. During the the scanning process we scanned the part shown from \textbf{multiple} directions in order to capture as much as possible of the curved surface. The same part of the object is shown in Figure \ref{fig:sls_scan} but this time it was scanned using 3DUNDERWORLD-SLS. It should be noted that this is the results of a \textbf{single scan}. 

We applied hole-filling on both meshes with the same settings (i.e. of size less than 50 edges) and both meshes are shown with Gouraud shading. The mesh produced by the commercial solution contained 107888 faces formed by 54870 points, and the mesh produced by our solution contained 55068 faces formed by 28014 points. A possible explanation for the significant difference in the number of points and faces is the fact that the laser scanner works in "sweeps" therefore, as the scanner moves more points are added to the reconstructed result. When using SLS the same occurs when combining multiple scans. Although the number of points and number of faces of the high-end 3D scanner surpass those of the 3DUNDERWORLD-SLS it is evident that the 3DUNDERWORLD-SLS outperforms the commercial solution in the level of detail captured and the amount of information captured in a single scan. 

\section{Conclusion}
\label{sec:conclusion}
Although the theory behind the SLS systems is well documented and understood, there are still many issues one has to consider when developing or using SLS systems, which are currently lacking documentation. Many variants of SLS systems have already been proposed however, each one is tailored to a particular task. In this paper, we have presented all possible limitations, difficulties and solutions that one has to consider when involved in the design, development or use of SLS systems.

Furthermore, we have introduced the general-purpose, open-source 3DUNDERWORLD-SLS software and reported on the results of our extensive testing. Moreover, we have evaluated the proposed system on four different evaluation metrics and compared it to a high-end commercial 3D scanner. The produced results are of considerable high-fidelity and the system is currently being used as a documentation device in archaeological sites.

\section*{Acknowledgements}
3DUNDERWORLD-SLS v3.2 and prior versions, were supported by EC FP7 Marie Curie IRG-268256 - "Rapid Scanning and Automatic 3D Reconstruction of Underwater Sites" - 3DUNDERWORLD – http://www.3dunderworld.org. 

3DUNDERWORLD-SLS v4.0, was supported by Concordia University Strategic Hire Grant VH0003 - "Immersive and Interactive Visualizations of Realistic Virtual Environments".

The software and other documentation is publicaly available and can be downloaded from the ICT lab's \hyperlink{https://github.com/theICTlab/3DUNDERWORLD-Structured-Light-Scanner}{Github account}.

\section{APPENDIX: Implementation Details}
Structured-light scanning involves the acquisition and processing of a large size of data. This, coupled with the high computational complexity required to compute the 3D positions, real-time implementation is not yet feasible with these type of techniques. However, our implementation offers near real-time reconstructions primarily due to the following:
\begin{itemize}
\item A compact yet thread safe binary array to store the masks and decoded patterns.
\item
  Fast bit operations
\item
  Fast GPU implementation
\end{itemize}

\subsection{CPU Implementation Details}\label{implementation-details-on-cpu}
The CPU implementation of the SLS algorithm involves the following steps:
\begin{verbatim}
for each camera:
    generate Mask
    for each pixel which passes the mask test in camera:
        decode pixel and assign it into buckets
for each bucket:
    find corresponding pixels in both cameras
    undistort ray and find intersection points
    calculate intersection point as reconstructed point
\end{verbatim}

\subsubsection{Dynamic bit array}\label{dynamic-bit-array}

There are two versions of Dynamic bit arrays, one for the CPU implementation and another for the GPU implementation. They follow same
design theory -- Compact arrangement of binary bits.

8 bits of each byte are fully used to represent a mask or a pattern. Within a byte, the bit is arranged from lower to higher as shown below,

\[\text{High}~\underbrace{10011011}_{1~byte}~\text{Low}\]

All of the bits are stored in a array of unsigned chars
\texttt{vector\textless{}uchar\textgreater{}}. To operate on the bit
array, we first locate the byte where the target bit is located, and use bit operations on the char.

\paragraph{CPU Dynamic\_Bitset.h}\label{cpu-dynamicux5fbitset.h}

This bit set class encapsulates only one bit set in each object. Since we decode patterns once at a time on CPU. Bit sets can be thrown away when they have been encoded into an integer.

\subsection{GPU Implementation Details}\label{implementation-details-on-gpu}
The GPU implementation of the SLS algorithm involves the following steps, including certain kernel calls:

\begin{verbatim}
for each camera:
    compute masks and thresholds
    genPatternArray
    buildBucket
for each bucket:
    getPointCloud2Cam
\end{verbatim}

\begin{itemize}
\item
  genPatternArray -- This kernel decodes patterns into binary arrays.
  Each tread gose over the same pixel in the camera images, and generate a bit set array.
\item
  buildBucket -- This kernel inserts pixels into the buckets by their
  decoded pattern. Each thread decodes a camera pixel pattern and insert it into the corresponding bucket.
\item
  getPointCloud2Cam -- This kernel takes two buckets of cameras to reconstruct image. Each thread fetches pixels in the same bucket, then undistort and reconstruct the point.
\end{itemize}

As most of the steps are similar to the CPU implementation, the following sections will focus on the improvement on the arrangement of patterns using dynamic bit set on GPU and the parallel insertion of buckets. 

\subsubsection{GPU Dynamic\_Bitset}\label{gpu-dynamicux5fbitset}

The GPU implementation requires high concurrency and small device-host
communication. In this case, we put the patterns for all pixels in a
single bit array.

\begin{verbatim}
    unsigned char* bits;
    size_t BITS_PER_BYTE;
    size_t numElem;
    size_t bitsPerElem;
\end{verbatim}

Note that the number of bits are aligned to integer times of
\texttt{BITS\_PER\_BYTE}. Similar to the bit sets on CPU, each thread
will operate on the pattern of each pixel. Thus, there will be no more than two threads operated on the same memory, and race conditions are thus avoided. \[
\underbrace{\overbrace{\underbrace{\underbrace{00110010}_{byte}\underbrace{01110010}_{byte}}}^{\text{thread0}}_{\text{pattern for one pixel}}\cdots\underbrace{00110010}_{byte}\underbrace{01110010}_{byte}}_{\text{patterns for all pixels}}
\] 
For the example shown above, the patterns take 16 bits, i.e. 2 bytes, \emph{thread0} will only operate on the pattern for one pixel. Therefore, no threads will operate on the same bytes at the same time.

\subsubsection{Bit operations on Graycode
Generation}\label{bit-operations-on-graycode-generation}

For a binary array encoded in Gray code we use the following snippet to convert it to decimal which allows us to find the location of the projector's pixels.

\begin{lstlisting}[language=C]
 for (unsigned bit=1U<<31; bit>1; bit>>=1) {
    if (xDec & bit) xDec ^= bit >> 1;
 }
\end{lstlisting}

\subsubsection{GPU Generate Buckets}\label{gpu-generate-buckets}

Parallel implementation of buckets is a challenging task. As we insert camera pixels into buckets indexed by the projector's pixels with one thread processing one camera pixel, it is impossible to put the elements one after another since all threads are running parallel and one thread cannot determine if the insertion position is going to be used by another thread.

\begin{center}
\begin{figure}
    \centering
    \includegraphics{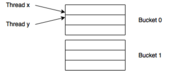}
    \caption{Inserting elements to the same bucket may trigger race condition.}
    \label{fig:race01}
\end{figure}
\end{center}

\begin{center}
\begin{figure}
    \centering
    \includegraphics{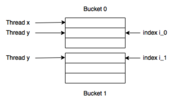}
    \caption{Each bucket will maintain an atomicly increasing index indicates where to insert next element within the bucket.}
    \label{fig:race02}
\end{figure}
\end{center}

In the Figure \ref{fig:race01}, the race condition occurs when \texttt{thread\ x}
and \texttt{thread\ y} both want to insert into bucket 0. The solution is to pre-allocate the maximum possible memory for each bucket and
use an atomically increasing index to indicate where to store the next element in the buckets, as depicted in Figure\ref{fig:race02}.

The atomic operation will only be triggered when two threads are trying to insert to the same bucket, thus the serialization is low and the impact on
performance is not obvious.

\subsection{Benchmark}\label{benchmark}
A comparison between the CPU implementation of v4 and the previous versions show a speed increase of at least two orders of magnitude. The configuration of the machine is given in the
following table

\begin{center}
\begin{tabular}{cl}
\hline
CPU     &   Intel i7 4771k\\
Memory  &   16GB    \\
GPU     &   nVidia GTX 770  \\
vMemory &   2GB \\
\hline
\end{tabular}
\end{center}

A comparison between the GPU and CPU of v4 on the same configuration with the Alexander dataset is shown in the following table.

\begin{center}
\begin{tabular}{cl}
\hline
GPU     &   12.940s\\
CPU     &   29.67s\\
\hline
\end{tabular}
\end{center}

\subsection{Code Structure}\label{code-structure}

The latest version of the 3DUNDERWORLD-SLS (v4) was designed with extensibility and maintainability in mind. 3DUNDERWORLD-SLS is itself a library that can be easily integrated into other applications using CMake building system. As it is shown below,
\texttt{lib} folder contains all of the implementation of library and
\texttt{app} folder contains some test applications using the libraries
as references.

\begin{verbatim}
src
├── app
│   ├── App_Calib.cpp
│   ├── App.cpp
│   ├── App_CUDA.cu
│   ├── App_Graycode.cpp
│   └── CMakeLists.txt
└── lib
    ├── calibration
    │   ├── Calibrator.cpp
    │   ├── Calibrator.hpp
    │   └── CMakeLists.txt
    ├── core
    │   ├── Camera.cpp
    │   ├── Camera.h
    │   ├── CMakeLists.txt
    │   ├── Dynamic_Bitset.cpp
    │   ├── Dynamic_Bitset.h
    │   ├── fileReader.cpp
    │   ├── fileReader.h
    │   ├── log.cpp
    │   ├── log.hpp
    │   ├── Projector.h
    │   ├── Ray.h
    │   ├── Reconstructor.cpp
    │   ├── ReconstructorCPU.cpp
    │   ├── ReconstructorCPU.h
    │   └── Reconstructor.h
    ├── GrayCode
    │   ├── CMakeLists.txt
    │   ├── GrayCode.cpp
    │   └── GrayCode.hpp
    └── ReconstructorCUDA
        ├── CMakeLists.txt
        ├── CUDA_Error.cuh
        ├── Dynamic_bits.cu
        ├── Dynamic_bits.cuh
        ├── fileReaderCUDA.cu
        ├── fileReaderCUDA.cuh
        ├── ReconstructorCUDA.cu
        └── ReconstructorCUDA.cuh
\end{verbatim}

\subsubsection{core}\label{core}

As the name indicates, core library contains the base classes of reconstructor, camera, and patterns. Please note that a CPU reconstructor also locates here since all of the machines should be able
to run it.

\subsubsection{ReconstructorCUDA}\label{reconstructorcuda}

\texttt{ReconstructorCUDA} class is an implementation of reconstruction
on GPU that inherited from reconstructor in the \texttt{core}. A GPU
version of Dynamic\_bits and fileReader are also implemented to be
compatible with device code on GPU.

\subsubsection{GrayCode}\label{graycode}

This library is used to generate gray code pattern. Only a small change
has bee applied from the original code.

\subsubsection{Calibration}\label{calibration}

Camera Calibration with OpenCV.

\bibliographystyle{elsarticle-num}
\bibliography{acmsmall-sample-bibfile}

\end{document}